\documentclass{bmvc2k}

\usepackage{microtype}
\usepackage{enumitem}
\usepackage[export]{adjustbox}

\title{Temporally Coherent Person Matting\\Trained on Fake-Motion Dataset}

\addauthor{Ivan Molodetskikh}{ivan.molodetskikh@graphics.cs.msu.ru}{1}
\addauthor{Mikhail Erofeev}{merofeev@graphics.cs.msu.ru}{1}
\addauthor{Andrey Moskalenko}{andrey.moskalenko@graphics.cs.msu.ru}{1}
\addauthor{Dmitry Vatolin}{dmitriy@graphics.cs.msu.ru}{1}

\addinstitution{
 Lomonosov Moscow State University\\
 Moscow, Russia
}

\runninghead{Molodetskikh \etal}{Temporally Coherent Person Matting}

\def\etal{\emph{et al}\bmvaOneDot}

\makeatletter
\def\blfootnote{\gdef\@thefnmark{}\@footnotetext}
\makeatother

\setlist{nosep}

\begin{document}

\maketitle

\begin{abstract}
We propose a novel neural-network-based method to perform matting of videos depicting people that does not require additional user input such as trimaps. Our architecture achieves temporal stability of the resulting alpha mattes by using motion-estimation-based smoothing of image-segmentation algorithm outputs, combined with convolutional-LSTM modules on U-Net skip connections.

We also propose a fake-motion algorithm that generates training clips for the video-matting network given photos with ground-truth alpha mattes and background videos. We apply random motion to photos and their mattes to simulate movement one would find in real videos and composite the result with the background clips. It lets us train a deep neural network operating on videos in an absence of a large annotated video dataset and provides ground-truth training-clip foreground optical flow for use in loss functions.
\end{abstract}

\section{Introduction}

Matting, or finding the opacity map of a foreground object, is a crucial image- and video-processing operation. The goal is to determine the foreground object's pixel-transparency values, as well as the true foreground and background pixel-color values. Matting yields a cutout image of an object that can be overlaid on an arbitrary background using composition.

Background replacement and, thus, foreground extraction are popular in amateur filming, videoconferencing and live streaming. They are also useful for augmented reality. For these areas it's desirable to avoid the need for a special background stage or limitations on the foreground-object color imposed by the traditional chroma-keying method.

Matting helps solve these problems, as it doesn't require special backgrounds or equipment. But these characteristics make it a considerably more difficult task than chroma keying because in general, it is impossible to tell which objects are considered to be in the foreground and thus require extraction.

Foreground-object selection usually employs trimaps, which annotate every input-image pixel to indicate whether it's foreground, background or unspecified. The matting algorithm must then find the transparency of the unspecified pixels. Other foreground-object-selection methods include scribbles and neural-network features. Manual foreground-object selection is sometimes impractical, especially for videos. We therefore tackle the problem of video matting that considers a fixed class---people---to be the foreground. In particular, we investigate a case in which a single person or just a few people are in the foreground. We use an image-segmentation method to generate a probability map of each video pixel belonging to a person and then use this map to guide our neural-network matting algorithm.

Another video-matting challenge is maintaining temporal coherence in the resulting alpha mattes. Notably, even when trimaps are available for every frame in a video, frame-by-frame application of most image-matting approaches yields alpha mattes that visibly vibrate and flicker when converted back to video---an unpleasant and distracting artifact. We introduce a temporal-smoothing algorithm for image-segmentation probability maps and use convolutional LSTM layers to increase the temporal stability of the result.

As of this writing, no publicly available video dataset containing ground-truth alpha mattes of people is large enough to train a deep neural network. There are, however, large datasets of human portraits with ground-truth alpha mattes. We propose a novel fake-motion algorithm to generate training video clips from image foregrounds and video backgrounds by distorting the foreground image throughout the clip. We show that fake motion can be used to train a video-matting deep neural network capable of processing real input videos to produce temporally stable alpha mattes.

Our main contributions in this paper are as follows:
\begin{itemize}
   \item A novel deep-neural-network method for matting videos in which people are foreground objects, with no additional user input requirements.
   \item A fake-motion algorithm for generating neural-network training video clips from a dataset of images with ground-truth alpha mattes and background videos.
   \item A motion-estimation-based method for temporal smoothing of the image-segmentation method's probability output, considerably improving the output's temporal stability.
\end{itemize}

\section{Related Work}

In this section we provide an overview of existing image- and video-matting techniques, focusing on semi- and fully automatic semantic methods.

\subsection{Image Matting}

Image matting, a subtask of video matting, has been an active research topic in recent years. This situation likely owes to the increasing popularity of neural networks: matting is one problem where neural-network-based methods have exhibited outstanding results compared with classical ones. Here we review the latest works that are most relevant to our proposed approach; a comprehensive overview of traditional algorithms appears in~\cite{wang2008image} and~\cite{li2019survey}.

The first algorithm for automatic human-portrait matting without additional input data was~\cite{shen2016deep}. Shen \etal created a dataset containing 2,000 human portraits with ground-truth alpha mattes, and they trained a neural network to predict the trimap given an image. A differentiable classical matting algorithm then processes the image together with the trimap to find the pixel-transparency values in unknown regions, enabling end-to-end network training. The neural network also receives an ``average'' transparency map over the training dataset, aligned with the input photo, to reduce the number of falsely labeled foreground pixels.

Chen \etal~\cite{chen2018semantic} used two neural networks to improve the quality of automatic portrait matting. The first predicts the trimap, and the second uses that result to aid in predicting the transparency map. The method then combines the transparency map with the trimap to obtain the resulting map. The networks first undergo training separately, then the system undergoes end-to-end training. Another chief contribution of this work is a dataset containing 34,425 human portraits with ground-truth alpha mattes~\cite{aisegmentcom_dataset}.

Seo \etal~\cite{seo2019towards} explore an approach targeting real-time execution on mobile devices. Their network is lightweight, using depthwise-separable convolutions and weight quantization. A limitation is low matte quality, especially for high-resolution input photos.

The method of Zhang \etal~\cite{Zhang_2019_CVPR} can automatically produce alpha mattes for images with various foreground objects. It uses a neural network comprising three parts: foreground and background probability-prediction modules and a fusion network that outputs a map of blending coefficients for computing the final transparency map.

Liu \etal~\cite{Liu_2020_CVPR} proposed an algorithm consisting of multiple neural networks, allowing them to use both coarse and fine portrait annotations; the former are easy to obtain, but the latter require more effort to create. First, a mask-prediction network outputs a coarse semantic mask. A quality-unification network then increases the mask quality to make it consistent with high-quality annotations. Finally, a matting-refinement network uses the improved mask as a guide to produce the final alpha matte.

\subsection{Video Matting}

Video matting has seen less active research than image matting, presumably because providing dense trimap data is usually impractical and because employing sparse or no additional input is much more difficult. Further complications arise owing to the large size of video data as well as the lack of big, publicly available datasets with video-matting ground truth.

Backes and Oliveira~\cite{backes2019patchmatch} described a method that produces temporally coherent mattes for intermediate video frames given mattes for the shot's first frame and last frame. It propagates the mattes from both ends using optical flow computed by a PatchMatch-based algorithm.

Zou \etal~\cite{zou2019unsupervised} proposed a traditional approach that represents background and foreground pixels using multiplication of a dictionary matrix and a sparse code matrix. An optimization process trains the dictionary on the input video, taking temporal stability into account. This approach requires a sparse-trimap input for the first frame of each shot.

Shen \etal~\cite{shen2017automatic} presented a portrait-video-matting technique that requires no additional trimap or scribble input. It does, however, use manually provided video-background sample images to help the neural network handle complex backgrounds.

Sengupta \etal~\cite{Sengupta_2020_CVPR} proposed a semiautomatic matting approach where the user, instead of handcrafting a trimap, need only supply a photo of the background. Exact alignment of the target video and background photo is unnecessary. Like our proposed approach, this algorithm takes as input a processed output of an image-segmentation network. One limitation involves nonstatic backgrounds, such as a waterfall or extensive camera motion. In these cases the algorithm can falsely mark some background elements as foreground.

Oh \etal~\cite{oh2018fast} proposed a video object segmentation method that uses synthetic training sample generation by applying random geometric transformations and color perturbation both to complete images with object masks and to separate foreground and background images. They further improved this method in~\cite{oh2019video} by generating short video clips in a similar fashion. Our approach builds on the same principles and includes a novel fake-motion component that further improves the quality of the generated training clips. 

\section{Proposed Approach}

\begin{figure*}
   \begin{center}
      \includegraphics[width=\linewidth]{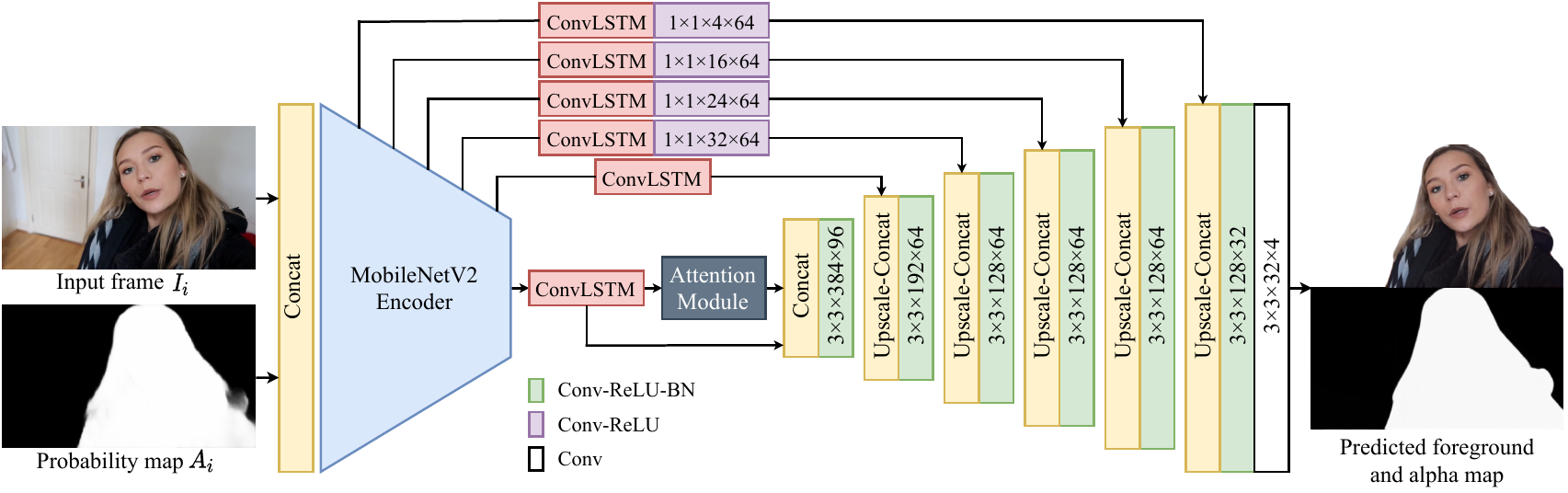}
   \end{center}
   \caption{Architecture of the proposed neural network.}%
   \label{fig:architecture}
\end{figure*}

Our approach uses a fully convolutional neural network to predict the alpha matte. Figure~\ref{fig:architecture} shows an overview. The network takes as input the current video frame $I_i$ and a coarse person-probability map $A_i$.

\subsection{Architecture}

The neural network follows the U-Net architecture with MobileNetV2~\cite{Sandler_2018_CVPR} as the encoder's backbone. We take the outputs of the 2nd, 4th, 7th, 14th and 18th blocks of the TorchVision~\cite{torchvision} implementation for skip connections and increase the first layer's channel count from three to four in order to accommodate the extra probability-map input. Skip connections go through convolutional LSTM modules~\cite{convlstm}, which provide temporal coherence to the output alpha mattes.

Additionally, we use an attention module on the bottleneck connection. First, we concatenate the bottleneck feature map with three index maps containing each pixel's training-clip frame number, x-coordinate and y-coordinate. Then we use two fully connected layers to reduce the number of dimensions in the flattened feature maps from 323 to 64 and to generate query and key maps. We similarly employ a fully connected layer to generate the value map, but we employ the 320-dimensional flattened bottleneck feature map as is without appending the index maps. Next, we apply scaled dot-product attention as Vaswani \etal described~\cite{vaswani2017attention} and reshape the result to restore its temporal and spatial dimensions.

To generate the coarse person-probability map $A_i$, we use a pretrained DeepLabv3+~\cite{deeplabv3plus2018} image-segmentation network from TorchVision. We bilinearly resize the input image to 520\texttimes 520 resolution, process it with DeepLabv3+, take a channel-wise softmax of the output and extract the channel corresponding to the human class. The final step is to return this map to the original input-image resolution using another bilinear resize. DeepLabv3+ is part of our proposed method, and we use it during both training and inference.

\subsection{Clip Generation}

\begin{figure}[t]
   \begin{center}
      \begin{tabular}{rl}
        (a) & \includegraphics[width=.5\linewidth,valign=m]{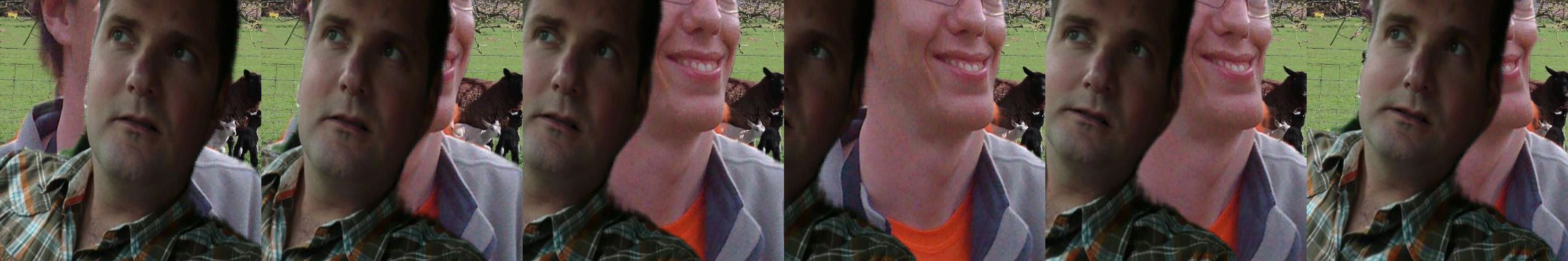}\\
        (b) & \includegraphics[width=.5\linewidth,valign=m]{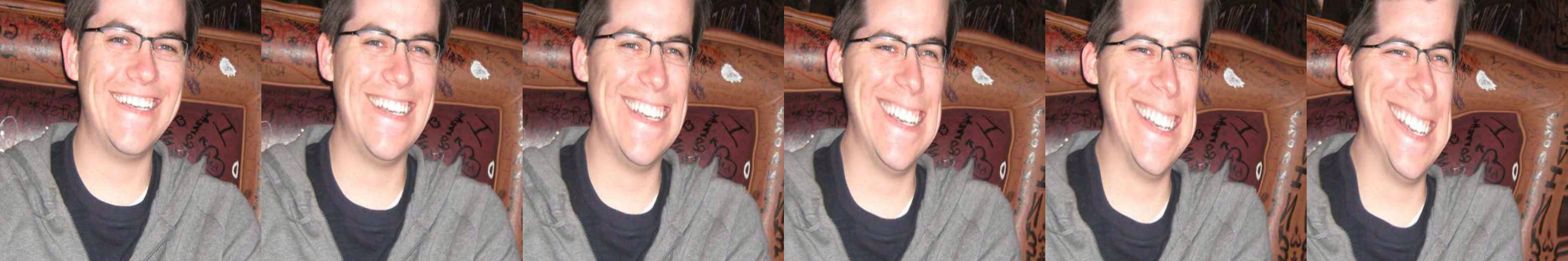}\\
        (c) & \includegraphics[width=.5\linewidth,valign=m]{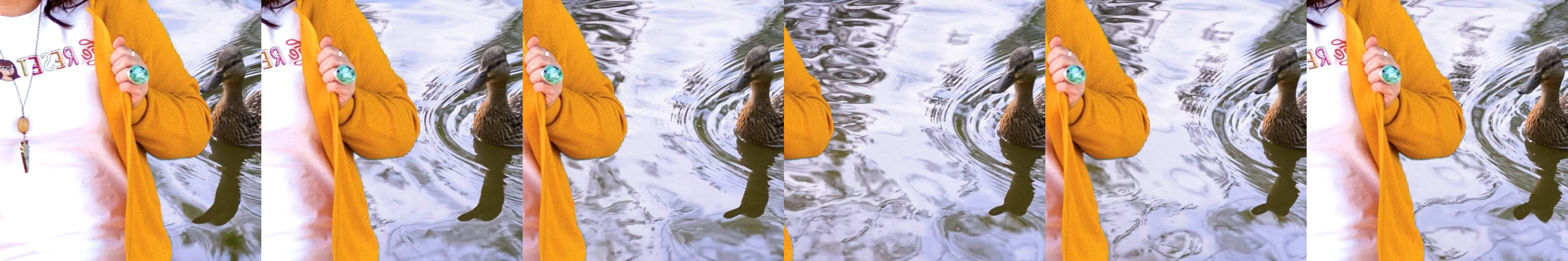}
       \end{tabular}
   \end{center}
   \caption{Examples of generated clips: (a) final training clip; (b) clip from the base fake-motion algorithm; (c) example of the fake-motion shift component's effect.}%
   \label{fig:training-clips}
\end{figure}

Training a deep neural network for video matting requires a large dataset containing videos with ground-truth alpha mattes. Unfortunately, no such dataset is publicly available. There are, however, large datasets of human portraits with ground-truth alpha mattes~\cite{aisegmentcom_dataset,shen2016deep}. In this section we describe an algorithm for generating training video clips using annotated foreground images and background videos. Figure~\ref{fig:training-clips}~(a) shows an example.

To generate an $N$-frame-long training clip, we first select a random background clip of length $N$ and two random foreground portraits. The fake-motion procedure, described below, processes the portraits to yield two foreground clips of length $N$ along with two ground-truth optical-flow clips. In half of the cases, we composite the foregrounds and the background; for the others, we employ the first foreground clip as is. Using two foreground clips at once improves the network performance for videos featuring multiple people in the frame. Additionally, in 5\% of the cases, we leave the background clip as is, without any compositing, to better handle video areas that lack any people. Finally, we apply JPEG compression with a random quality between 30\% and 80\% to the resulting clips.

The 50\% probability of omitting the compositing, along with the final JPEG compression, prevents network overfitting to fake composites: the resulting clips are not composited in half of the cases, and even when they are, JPEG-compression artifacts make it more difficult for the network to ``find'' the composition edges.

\subsection{Fake Motion}
\label{sec:fake-motion}

The fake-motion procedure generates random optical-flow maps at three scales. First, it produces one optical-flow vector by drawing two values ($x$ and $y$) from the normal distribution $\mathcal{N} \left( 0, 32 \right)$. The vector can be treated as a $1\times1$ optical-flow-vector map. This output corresponds to strong global motion. The procedure then generates a $\lfloor\frac{H}{128}\rfloor\times\lfloor\frac{W}{128}\rfloor$ optical-flow-vector map by drawing values from $\mathcal{N} \left( 0, 16 \right)$. This output represents a finer motion component. Finally, the procedure generates a $\lfloor\frac{H}{32}\rfloor\times\lfloor\frac{W}{32}\rfloor$ map using $\mathcal{N} \left( 0, 4 \right)$. This output is the finest motion component. We upscale these three maps to the input-image size ($H\times W$ pixels), sum them and then divide them by the target-clip frame count, $N$, to yield the per-frame optical flow. Figure~\ref{fig:training-clips}~(b) shows an example clip generated using this algorithm.

To improve network resilience when a person partially leaves the video frame, we added to the optical flow a component that over the clip's duration shifts the person halfway out of a frame and back, with a probability of $\frac13$. The shift is random under the condition that no opaque pixels touch the side opposite to the shift: this condition prevents border-replication artifacts involving part of a person. Figure~\ref{fig:training-clips}~(c) shows the shift component's effect. Additionally, since we employ two foreground clips for each training clip, our approach adds a second component that applies a half-frame initial shift to the foreground. This step reduces the chance that two foreground portraits completely overlap each other.

We use the resulting optical-flow maps to warp the input foreground image and alpha mask to produce the foreground clip with the desired frame count.

\subsection{Loss Functions}

\begin{figure}[t]
   \begin{center}
      \begin{tabular}{rl}
      $\alpha'_i$                          & \includegraphics[width=0.4\linewidth,valign=m]{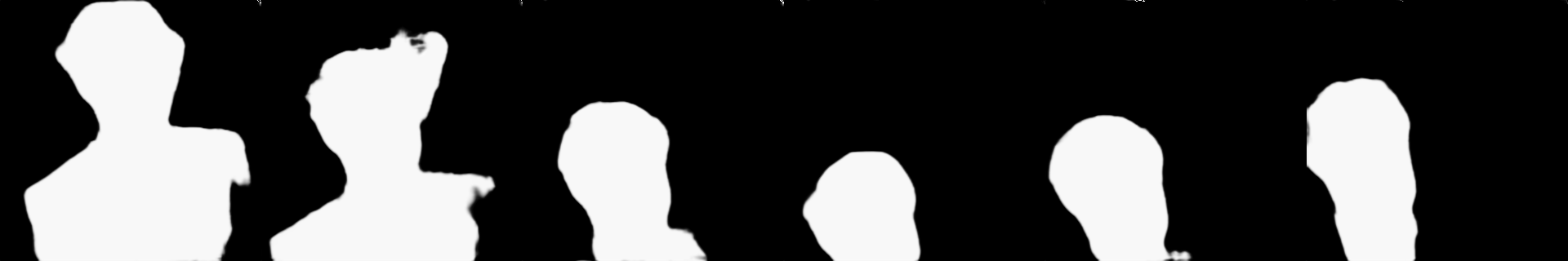}\\
      $\alpha_1^{\prime\rightarrow i}$     & \includegraphics[width=0.4\linewidth,valign=m]{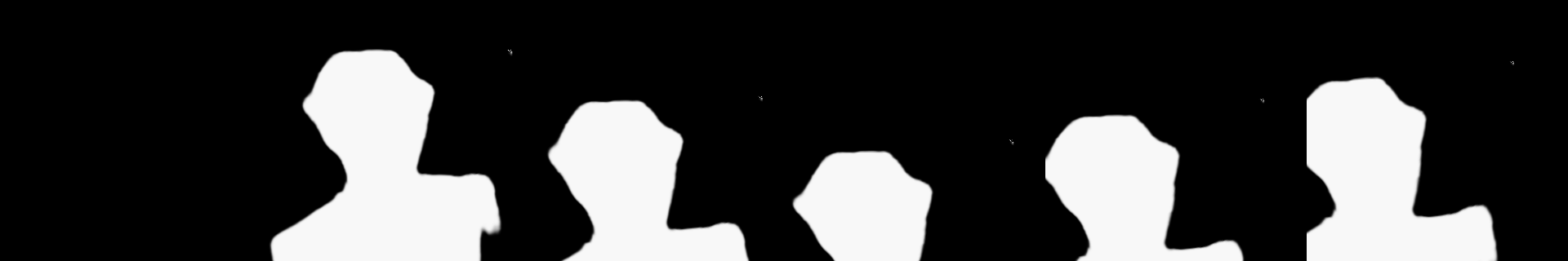}\\
      $V_1^{\rightarrow i}$                & \includegraphics[width=0.4\linewidth,valign=m]{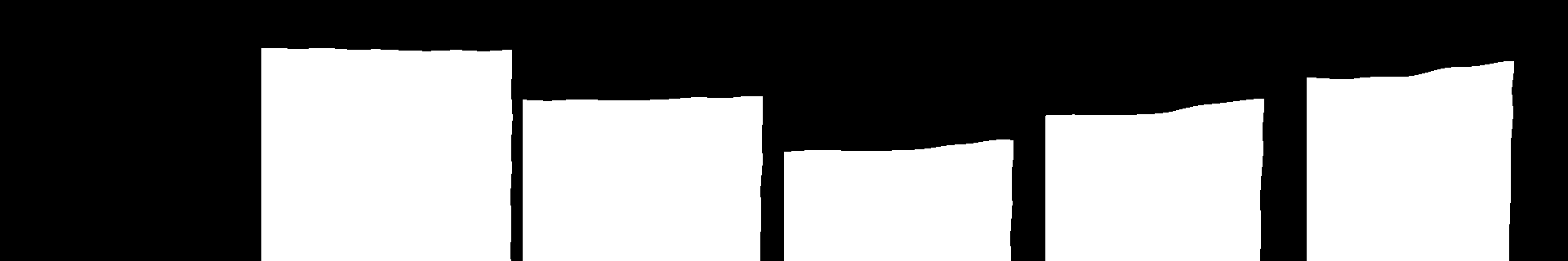}\\
      $\alpha_{i-1}^{\prime\rightarrow i}$ & \includegraphics[width=0.4\linewidth,valign=m]{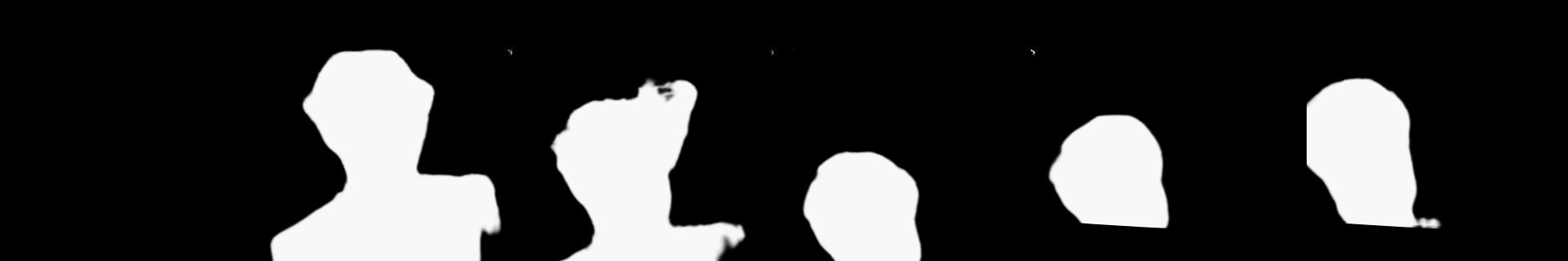}\\
      $V_{i-1}^{\rightarrow i}$            & \includegraphics[width=0.4\linewidth,valign=m]{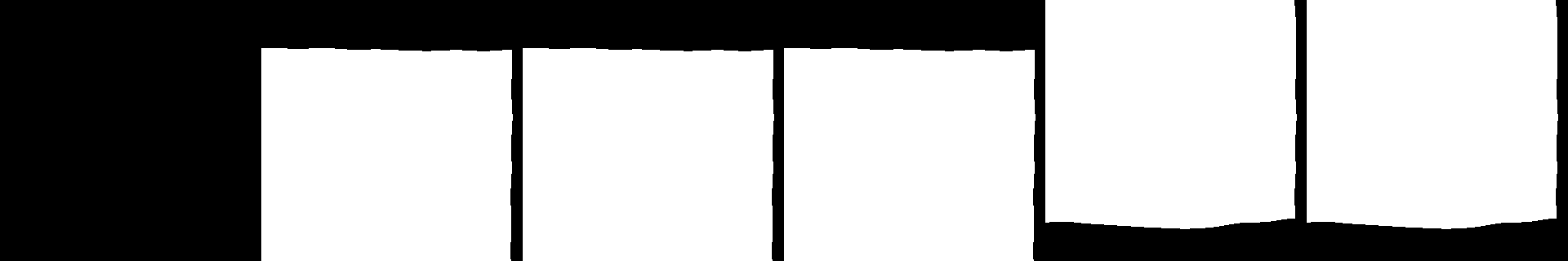}
      \end{tabular}
   \end{center}
   \caption{Examples of maps for $L_G$ and $L_L$ computation.}%
   \label{fig:loss-maps}
\end{figure}

Training of our network uses a four-component loss function: \(L = L_\alpha + L_G + L_L + \frac14 L_F\). The first component is a pixel-wise $L_1$ distance between the predicted alpha map $\alpha'_i$ and ground-truth alpha map $\alpha_i$:
\(
L_\alpha = \sum_{i=1}^N \sum_{j=1}^{H} \sum_{k=1}^{W} \left| \alpha'_{ijk} - \alpha_{ijk} \right| / \left(N\cdot H\cdot W\right),
\)
where $N$ is the number of frames in the training clip and $H$ and $W$ are the clip's pixel height and width, respectively.

The second and third components ensure the result is temporally stable. $L_G$ compares the first frame's alpha map with each subsequent frame's alpha map, while $L_L$ compares alpha maps between successive frames:
\begin{equation}
    L_G = \frac{\sum_{i=2}^N \sum_{j=1}^{H} \sum_{k=1}^{W} \left| \alpha'_{ijk} - \alpha_{1jk}^{\prime\rightarrow i} \right| V_{1jk}^{\rightarrow i}}{\left(N-1\right)\cdot H\cdot W},\quad L_L = \frac{\sum_{i=2}^N \sum_{j=1}^{H} \sum_{k=1}^{W} \left| \alpha'_{ijk} - \alpha_{i-1jk}^{\prime\rightarrow i} \right| V_{i-1jk}^{\rightarrow i}}{\left(N-1\right)\cdot H\cdot W}.
\end{equation}
Here, $\alpha_l^{\prime\rightarrow i}$ denotes the predicted alpha map for the $l$th frame, warped to the $i$th frame according to the ground-truth optical flow. Also, $V_l^{\rightarrow i}$ is a mask of pixels for which the comparison is valid: it equals 1 for pixels of the $i$th frame that, according to the optical flow, correspond to pixels of the $l$th frame, and it equals 0 for pixels of the $i$th frame that correspond to pixels beyond the boundaries of the $l$th frame.  Figure~\ref{fig:loss-maps} shows examples of these maps.

The final component, $L_F$, is a pixel-wise $L_1$ distance between the predicted RGB foreground $\mathit{FG}'_i$ and the ground-truth foreground $\mathit{FG}_i$, multiplied by the predicted alpha map:
\begin{equation}
   L_F = \frac{\sum_{i=1}^N \sum_{j=1}^{H} \sum_{k=1}^{W} \alpha'_{ijk} \sum_{c=1}^3 \left| \mathit{FG}'_{ijkc} - \mathit{FG}_{ijkc} \right| }{N\cdot H\cdot W\cdot 3}.
\end{equation}

\subsection{Dataset}

We used a dataset consisting of 27 background videos from DAVIS~\cite{Perazzi2016}. They contain no humans and are 80 frames long on average. We also used 40,074 foreground portraits with ground-truth alpha mattes: 34,425 images from~\cite{aisegmentcom_dataset}, 2,000 images from~\cite{shen2016deep} and 3,649 images we annotated ourselves. We set aside 4,000 foreground images for validation and used the remaining 36,074 for training.

Additionally, we used the following augmentations: random rotation of up to \textpm 15\textdegree, crop down to 20\%, horizontal flip and brightness-contrast adjustment on foregrounds; random crop down to 8\% of the frame size, horizontal flip and brightness-contrast adjustment on backgrounds; and JPEG compression of the composited clips at 30--80\% quality. These augmentations, together with shifts mentioned in Section~\ref{sec:fake-motion}, yield training-clip generation similar to that in~\cite{oh2018fast,oh2019video}. The main fake-motion component then builds on top of this procedure to further improve quality. 

\subsection{Training Setup}

\begin{figure}[t]
   \begin{center}
      (a) \raisebox{-2cm}{\includegraphics[width=.43\linewidth]{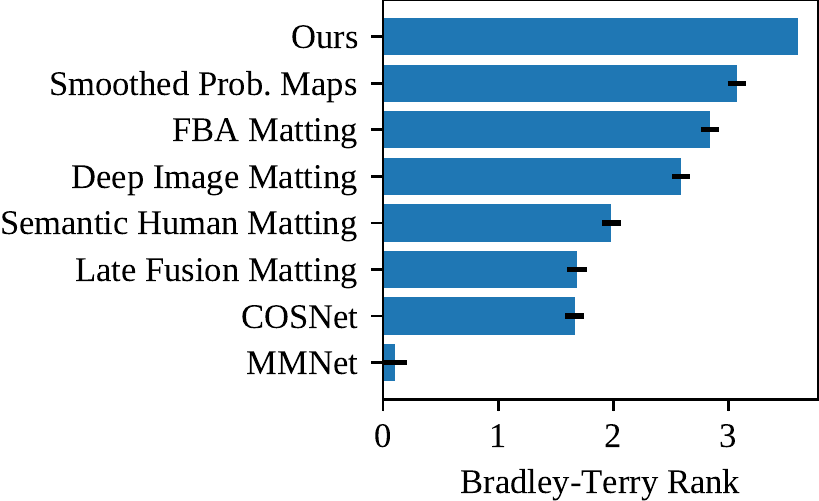}} (b) \raisebox{-2cm}{\includegraphics[width=.43\linewidth]{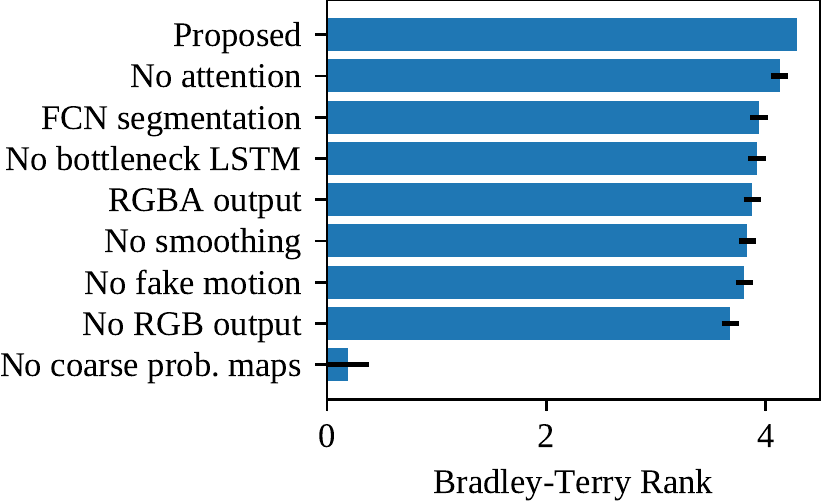}}
   \end{center}
   \caption{Subjective evaluation results compared to (a) other methods; (b) network variations. Error bars show 95\% confidence intervals for differences from our method's rank.}%
   \label{fig:subjective}
\end{figure}

We implemented the model using PyTorch 1.2 and trained it on four NVIDIA Tesla P100 GPUs. Our batches contained eight clips, each comprising six frames at 520\texttimes 520 resolution. We used the Adam optimizer with a learning rate of 0.001, decreasing the rate by a factor of 0.7 after each epoch. Our test models generally converged after seven epochs, or about 9.5 hours of training.

\subsection{Inference}
\label{sec:inference}

Inference only uses the transparency-map output, ignoring the RGB foreground output because we found it frequently contains color artifacts.

To better handle sudden fast motion, we apply the model to the target video in four-frame windows. In addition, applying temporal smoothing to the person-probability maps, provided in $A_i$, boosts the temporal stability of the final output. We use a motion-estimation-based algorithm, described below, to temporally smooth the probability maps.

Our proposed method can process a 1280\texttimes 720 video at about 1.7 frames per second on a single NVIDIA TITAN Xp GPU and an Intel Xeon E5-2683 v3 CPU running at 2 GHz.

\subsubsection{Optical-Flow and Consistency Maps}
First, a block-based motion-estimation algorithm from~\cite{grishin2008video3694423} processes pairs of subsequent frames from the input video $\left\{I_i\right\}^N_{i=1}$ in forward and reverse order to yield forward and backward motion vectors in the form of optical-flow maps: $\left\{F_i\right\}^N_{i=2}$ and $\left\{B_i\right\}^{N-1}_{i=1}$. Next, we warp the backward-optical-flow maps in accordance with the forward-optical-flow maps, denoting the resulting maps as $\left\{B^{\rightarrow i}_{i-1}\right\}^N_{i=2}$. We then subtract the warped backward-optical-flow maps from the forward-optical-flow maps: $D_i = F_i - B^{\rightarrow i}_{i-1}$. Our approach computes an $L_2$ norm for every pixel in the resulting maps: $L_{ijk} = \sqrt{D^2_{ijkx} + D^2_{ijky}}$, $i = \overline{2, N},\ j=\overline{1,H},\ k=\overline{1,W}$, where $x$ and $y$ are the optical-flow-vector components. Finally, we exponentiate the resulting maps to get consistency maps: $C_i = \exp \frac{-L_i}{100}$. Consistency maps have values close to 1 when the corresponding forward- and backward-optical-flow vectors are similar (indicating the vectors are likely correct) and values close to 0 otherwise.

\begin{table}
\begin{center}
\setlength{\tabcolsep}{5pt}
\begin{tabular}{r|rrr|rrr}
   \hline
   Method & SSDA & dtSSD & MESSDdt & SSDA & dtSSD & MESSDdt \\
   \hline
   & & city & & \multicolumn{3}{c}{snow} \\
   \hline
   Ours & \textit{69.651} & \textbf{15.314} & \textbf{0.695} & \textit{56.338} & \textit{30.240} & \textit{0.662} \\
   Smoothed Prob. Maps & 91.577 & \underline{17.609} & \underline{1.291} & 65.772 & 35.133 & 1.264 \\
   FBA Matting & \underline{57.700} & \textit{30.825} & \textit{1.613} & \underline{27.113} & \underline{20.881} & \underline{0.423} \\
   Deep Image Matting & 97.506 & 47.107 & 3.258 & 59.648 & 41.463 & 2.128 \\
   Sem. Human Matting & 108.393 & 53.086 & 5.696 & 71.844 & 43.689 & 2.595 \\
   Late Fusion Matting & \textbf{44.766} & 31.621 & 4.152 & \textbf{24.602} & \textbf{19.484} & \textbf{0.341} \\
   COSNet & 271.878 & 62.798 & 22.387 & 156.617 & 58.536 & 9.424 \\
   MMNet & 154.656 & 62.439 & 13.580 & 347.065 & 143.696 & 58.429 \\
   \hline
\end{tabular}
\end{center}
\caption{Objective evaluation results on \url{videomatting.com} clips. The best result is shown in \textbf{bold}, the second-best is \underline{underlined} and the third-best is shown in \textit{italics}.}
\label{tab:objective}
\end{table}

\subsubsection{Probability-Map Smoothing}
In inference, the optical-flow and consistency maps aid in smoothing the person-probability maps $\left\{p_i\right\}^N_{i=1}$ from DeepLabv3+. First, we compute the pixel-wise confidence maps, $s_i = 4\left(p_i - \frac12\right)$, then the smoothed person-probability maps:
\begin{equation}
    A_1 = p_1,\quad    A_i = s_i \cdot p_i + \left( 1 - s_i \right) \left( C_i \cdot A_{i-1}^{\rightarrow i} \right),\quad i=\overline{2,N},
\end{equation}
where $A_{i-1}^{\rightarrow i}$ is the probability map $A_{i-1}$ warped in accordance with the optical-flow map $F_i$.

Intuitively, for pixels with a person probability close to 0 or 1 (the image-segmentation network is ``certain'' about its answer), this probability is the final value; for pixels with person probability close to $\frac12$, the final value is from the previous frame, along the optical flow, scaled in accordance with the motion-vector consistency.

Temporal smoothing only occurs during the inference step, because applying it during training failed to improve the results.

\section{Experiments}

We conducted subjective and objective evaluations of our approach in comparison with the following methods:
\begin{itemize}
    \item Deep Image Matting~\cite{Xu_2017_CVPR}: our own implementation, trained on the Deep Image Matting dataset.
    
    \item FBA Matting~\cite{forte2020f}: official implementation and model~\cite{fba_matting_impl}, trained on the Deep Image Matting dataset.
    
    \item COSNet~\cite{Lu_2019_CVPR}: official implementation and model~\cite{cosnet_impl}, trained on the DAVIS16 dataset.
    
    \item MMNet~\cite{seo2019towards}: official implementation~\cite{mmnet_impl}, trained by us on the AISegment dataset~\cite{aisegmentcom_dataset}.
    
    \item Semantic Human Matting~\cite{chen2018semantic}: unofficial implementation and model~\cite{semantic_human_matting_impl}, trained on the implementation author's private dataset.
    
    \item Late Fusion Matting~\cite{Zhang_2019_CVPR}: official implementation and model~\cite{late_fusion_matting_impl}, trained on the Deep Image Matting dataset and the authors' private dataset.
\end{itemize}

\begin{table}
\begin{center}
\setlength{\tabcolsep}{5pt}
\begin{tabular}{r|rrr}
   \hline
   Method & SSDA & dtSSD & MESSDdt \\
   \hline
   Ours & \textbf{102.371} & \textbf{72.880} & \textbf{1.818} \\
   Smoothed Prob. Maps & \underline{113.019} & \underline{79.253} & \underline{3.955} \\
   FBA Matting & \textit{114.101} & \textit{92.686} & \textit{4.613} \\
   Deep Image Matting & 128.205 & 113.675 & 7.693 \\
   Semantic Human Matting & 186.980 & 145.235 & 9.685 \\
   Late Fusion Matting & 454.597 & 222.736 & 69.992 \\
   COSNet & 610.056 & 142.895 & 33.037 \\
   MMNet & 222.333 & 124.414 & 15.006 \\
   \hline
\end{tabular}
\end{center}
\caption{Objective evaluation results on 100 fake-motion clips. The best result is shown in \textbf{bold}, the second-best is \underline{underlined} and the third-best is shown in \textit{italics}.}
\label{tab:objective-fake-motion}
\end{table}

Deep Image Matting and FBA Matting are image-matting methods that have earned top ranks on the \href{https://videomatting.com}{\texttt{videomatting.com}} benchmark~\cite{Erofeev2015}, COSNet is an unsupervised video-segmentation technique that ranks high on the DAVIS16 benchmark~\cite{Perazzi2016}, and MMNet, Semantic Human Matting and Late Fusion Matting are automatic person-matting methods for images. We also used in our comparison person-probability maps $A_i$ after conducting the smoothing described in Section~\ref{sec:inference}, interpreting them as final alpha mattes.

To generate trimaps for~\cite{Xu_2017_CVPR} and~\cite{forte2020f}, we first generated a segmentation map using a pretrained DeepLabv3+ model~\cite{deeplabv3plus2018}. The next step was to dilate the person-class-segmentation map to get the unknown region and then erode the map to get the final foreground region, both with the number of iterations equal to 1\% of the image's pixel width.

\subsection{Subjective Evaluation}
\label{sec:subjective}

\begin{figure*}
   \begin{center}
      \includegraphics[width=\linewidth]{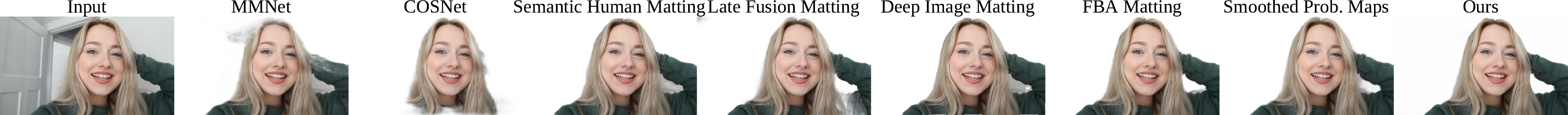}
      
      \includegraphics[width=\linewidth]{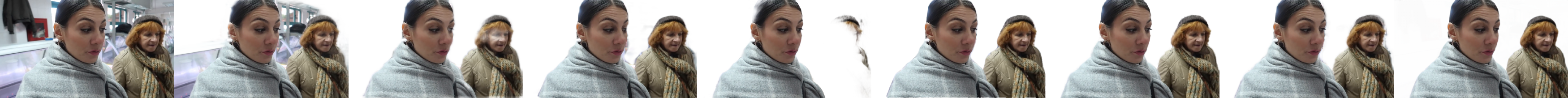}
   \end{center}
   \caption{Results of the evaluated methods on two of the test clips.}%
   \label{fig:results}
\end{figure*}

We conducted the subjective evaluation using \href{http://www.subjectify.us}{\texttt{subjectify.us}}. Participants were shown pairs of videos made by compositing test clips onto a static image background using alpha maps from the tested methods. They were informed that they would be comparing video-background-replacement methods and that their task for each video pair was to select the one they thought looked best. We used 2 test clips (city and snow) from~\cite{Erofeev2015} that fall in our target scope (videos containing people but no large semitransparent areas), 2 of our own clips and 55 additional clips from YouTube (371 frames long on average). Each participant viewed 12 video pairs, including 2 control pairs. Answers from participants who failed to select the best video in both control pairs were not counted. In total 25,202 pairwise selections were collected and used to fit a Bradley-Terry model~\cite{bradley1952rank}. Figure~\ref{fig:subjective}~(a) shows the results.

The image-matting methods~\cite{chen2018semantic,forte2020f,Xu_2017_CVPR,Zhang_2019_CVPR} did reasonably well, although temporal instability in the resulting mattes reduces viewer preference for these methods. Smoothed person-probability maps showed surprisingly good performance when interpreted as alpha maps. The main downside of this approach is coarse edge handling and the presence of alpha ``trails'' behind moving objects. COSNet~\cite{Lu_2019_CVPR}, despite being a video method, showed considerable flickering. MMNet~\cite{seo2019towards} exhibited the worst result, likely because it's unsuited to large images. In most comparisons, our proposed approach was preferred over the rest.

\subsection{Objective Evaluation}

Owing to the lack of applicable test videos with ground-truth alpha mattes, the objective evaluation is, unfortunately, limited. We performed the comparison using two test clips and three quality metrics from the \href{https://videomatting.com}{\texttt{videomatting.com}} benchmark. Table~\ref{tab:objective} shows the results. Our approach delivers either the best result or the third best after~\cite{Zhang_2019_CVPR} and~\cite{forte2020f}. This outcome likely owes to the metrics' limited ability to distinguish fine temporal coherence.

We additionally performed an objective evaluation on 100 clips generated using our fake-motion procedure. Table~\ref{tab:objective-fake-motion} shows the results. Our method performs better than the others---an expected result, given that we used similar clips for training.

\subsection{Ablation Study}

We performed a subjective comparison of several variations of our approach. They included using a different coarse-segmentation network (FCN~\cite{Long_2015_CVPR}) or omitting it entirely, using the full RGBA foreground output during inference rather than just using alpha, training the network without the RGB output, omitting the attention module or the convolutional LSTM module from the bottleneck connection, skipping the smoothing procedure described in Section~\ref{sec:inference}, and not using the main fake-motion component, described in the first paragraph of Section~\ref{sec:fake-motion}, during training clip generation.

We employed the same subjective-evaluation setup as in Section~\ref{sec:subjective}. Figure~\ref{fig:subjective}~(b) shows the results for 22,360 pairwise selections. Omitting the coarse segmentation has by far the most impact on perceived quality: the network has trouble coherently segmenting the person. Other modifications produce no drastic quality degradation, but how each component of our method positively contributes to the quality of the result is apparent. In particular, note that disabling the main fake-motion component while leaving all other standard geometric transformations of the foregrounds intact leads to considerable quality degradation in the resulting model.

\section{Conclusion}

We proposed a deep-neural-network method for person video matting with no additional user input. Our technique introduces a fake-motion algorithm for training-clip generation, allowing us to train the proposed neural network to produce temporally coherent results for real videos using a still-image foreground dataset. We also proposed a motion-estimation-based algorithm for making coarse image segmentation more temporally coherent, thereby improving the performance of our neural network.

Our approach performed better than alternatives in both subjective and objective evaluations. An ablation study demonstrated the impact of several of its main components. Figure~\ref{fig:results} shows results of the evaluated methods on two clips from the subjective study. More examples are provided in the supplementary video.
\blfootnote{Figures use frames from youtu.be/\{\href{https://youtu.be/OdWupUtuzAY}{OdWupUtuzAY}, \href{https://youtu.be/iVPTAcTQcN8}{iVPTAcTQcN8}, \href{https://youtu.be/NXW7hSmH8G4}{NXW7hSmH8G4}\}.}

\section{Acknowledgements}

This work was partially supported by Russian Foundation for Basic Research under Grant 19-01-00785 a and by Foundation for Assistance to Small Innovative Enterprises under Grant UMNIK 386GUCES8-D3/56342.

Model training has been conducted on the high-performance IBM Polus cluster of the CS MSU faculty: \url{http://hpc.cmc.msu.ru/polus}.

\bibliography{egbib}
\end{document}